\documentclass[sigconf]{acmart}

\usepackage[linesnumbered,ruled,vlined]{algorithm2e}
\usepackage{amsmath}
\usepackage{mathtools}
\usepackage{url} % Include the booktabs 
\usepackage{multirow} % Include the multirow 
\usepackage{amsfonts}
\usepackage{tabularx}
\usepackage{xcolor}

\usepackage[most]{tcolorbox}

\AtBeginDocument{%
  \providecommand\BibTeX{{%
    \normalfont B\kern-0.5em{\scshape i\kern-0.25em b}\kern-0.8em\TeX}}}

\begin{document}
\pagestyle{plain}
\pagenumbering{arabic}
\title{Multi-objective evolutionary GAN for tabular data synthesis}

\begin{comment}
\author{Author 1}
\email{Email}
\affiliation{%
  \institution{Institution}
  \city{City}
  \country{Country}
}

\author{Author 2}
\email{Email}
\affiliation{%
  \institution{Institution}
  \city{City}
  \country{Country}
}
\end{comment}
\author{Nian Ran}
\email{nian.ran@postgrad.manchester.ac.uk}
\affiliation{%
  \institution{The University of Manchester}
  \city{Manchester}
  \country{UK}
}

\author{Bahrul Ilmi Nasution}
\email{bahrul.nasution@postgrad.manchester.ac.uk}
\affiliation{%
  \institution{The University of Manchester}
  \streetaddress{School of Social Sciences}
  \city{Manchester}
  \country{UK}
}

\author{Claire Little}
\email{claire.little@manchester.ac.uk}
\affiliation{%
  \institution{The University of Manchester}
  \city{Manchester}
  \country{UK}
}

\author{Richard Allmendinger}
\email{richard.allmendinger@manchester.ac.uk}
\affiliation{%
  \institution{The University of Manchester}
  \streetaddress{Alliance Manchester Business School}
  \city{Manchester}
  \country{UK}
}

\author{Mark Elliot}
\email{mark.elliot@manchester.ac.uk}
\affiliation{%
  \institution{The University of Manchester}
  \streetaddress{School of Social Science}
  \city{Manchester}
  \country{UK}
}

\begin{abstract}
Synthetic data has a key role to play in data sharing by statistical agencies and other generators of statistical data products. Generative Adversarial Networks (GANs), typically applied to image synthesis, are also a promising method for tabular data synthesis. {\color{black}However, there are unique challenges in tabular data compared to images, eg tabular data may contain both continuous and discrete variables and conditional sampling, and, critically, the data should possess high utility and low disclosure risk (the risk of re-identifying a population unit or learning something new about them), providing an opportunity for multi-objective (MO) optimization.} Inspired by MO GANs for images, this paper proposes a smart MO evolutionary conditional tabular GAN (SMOE-CTGAN). This approach models conditional synthetic data by applying conditional vectors in training, and uses concepts from MO optimisation to balance disclosure risk against utility. Our results indicate that SMOE-CTGAN is able to discover synthetic datasets with different risk and utility levels for multiple national census datasets. We also find a sweet spot in the early stage of training where a competitive utility and extremely low risk are achieved, by using an Improvement Score. The full code can be downloaded from github\footnote{{\color{black}\url{https://github.com/HuskyNian/SMO\_EGAN\_pytorch}}}.
\end{abstract}

\maketitle

\section{Introduction}
Tabular data that originates from government organisations and other entities is a vital resource for researchers and therefore is disseminated for further evaluation and scrutiny among various organisations. During these sharing processes, statistical disclosure control (SDC) is used to protect the privacy of individuals represented in the dataset while striving to maintain optimal utility. With the growing sophistication of adversarial attacks seeking to re-identify data subjects and/or disclose information about them~\cite{intruders}, there is an increasing need for more frequent and robust SDC measures, which will often
result in reduced utility of the data~\cite{Purdam2007}.

Data synthesis techniques offer a more effective approach by learning from existing data and autonomously generating new tabular data that minimise disclosure risks. With rapid advances in machine learning, generative adversarial networks (GANs)~\cite{gan} have emerged as a potential solution for tabular data synthesis~\cite{ctgan,little2023federated,little2024benchmarking}. According to~\cite{framework}, conditional tabular GAN (CTGAN)~\cite{ctgan} shows some promise. However, when CTGAN methods have been compared with non-GAN-based methods, such as Synthpop~\cite{synthpop} and DataSynthesizer~\cite{data_synthesizer}, CTGAN is at a disadvantage in terms of utility and risk~\cite{tcap}.

These disadvantages are not surprising given that most GAN models depend on standard optimisation algorithms with a single objective function, which exacerbates vulnerability to instability~\cite{wgan} and mode collapse~\cite{Tomczak2022deep,arora2017generalization}. In addition, generating tabular data has its own challenges; in particular, varying types of data require different likelihood functions~\cite{ma2020vaem}. On the other hand, using variational approximation can lead to biases and difficulty in the derivation of the appropriate approximation~\cite{drefs2022evolutionary,nasution2022data}.

Recent studies have examined various approaches to overcome the problems, such as modifying architectures~\cite{makhzani2016adversarial,wang2023diffusiongan}, objective functions~\cite{wgan,nowozin2016fgan}, and the optimisation algorithms~\cite{saatci2017bayesian,turner2018metropolis,egan}. With various objective functions available, some researchers developed generative models that account for multiple conflicting objective functions using concepts from multi-objective optimisation~\cite{smoegan,albuquerque2019multiobjective}. Two techniques used for multi-objective optimisation in generative models are multiple gradient descent~\cite{albuquerque2019multiobjective} and evolutionary algorithms (EAs)~\cite{Nouri2023bievogan,egan,grantham2022deep,Zhou2023wgan}.

A recent advance in the GAN community is the application of evolutionary algorithms (EAs) to support training (described in more detail later), specifically the use of novel (also referred to as ``smart'' by the authors) reinforcement learning-based variation operators to produce new generators (offspring)~\cite{smoegan}. This approach was found to reduce the computational cost of the multi-objective evolutionary GAN and produced higher quality results. However, as with other existing evolutionary GAN studies, the implementation was limited to distributional and image data, such as MNIST, CIFAR-10, and LSUN bedroom~\cite{smoegan,albuquerque2019multiobjective,Nouri2023bievogan}. %{\color{black}However, there are unique challenges in tabular data compared to when dealing with images. For example, tabular data may contain both continuous and discrete variables, and be subject to conditional sampling. Importantly, synthetic tabular data should possess high utility and low disclosure risk, which is the risk of re-identifying a population unit or learning something new about them. The conflicting nature of utility and risk provides an opportunity for multi-objective (MO) optimization.} 

Consequently, the current paper proposes smart multi-objective evolutionary CTGANs (SMOE-CTGAN), which build on the foundation of SMOEGAN~\cite{smoegan} by incorporating multi-objective evolutionary algorithms into CTGAN to specifically enhance its performance with respect to utility and disclosure risk of tabular data. Furthermore, to better balance the trade off between utility and risk (i.e. discovery of effective generators at each training step), we propose a new metric, called the \textit{Improvement Score}, to be augmented onto the multi-objective optimiser, which comes after NSGA-II~\cite{nsga2} step in our case. Implementations and visualisations are available at github\footnotemark[1]. Having tested the model on four census datasets, our results indicate that the proposed model is able to generate synthetic datasets with both greater utility and lower risk compared to current state-of-the-art approaches~\cite{framework}. To our knowledge, this is the first study to implement SMOE within the context of CTGAN.

\begin{comment}
In summary, the contributions of this work are as follows:

\begin{itemize}
\item We introduce SMOE-CTGAN, a novel approach that successfully applies multi-objective optimisation in conjunction with deep reinforcement learning to conditional tabular generative adversarial networks, effectively utilising utility and disclosure risk as the multi-objectives. 
\begin {itemize}
\item Our model achieves higher utility scores and lower disclosure risks compared to the previous work and the CTGAN with our proposed Improvement Score. 
\end{itemize}

\item We analyze the merits of GANs in synthesising tabular data in terms of utility and risk, and propose a straightforward metric called Improvement Score, which can assess the relative superiority of one model over another.
\item We optimise the training process of our algorithm to decrease processing time without sacrificing effectiveness.

\end{itemize}
\end{comment}
The rest of the paper is structured as follows. Section~\ref{sec:back} describes the background of tabular GANs, EAs, and multi-objective optimisation. Section~\ref{sec:method} describes the proposed methodology including the new Improvement Score. Section~\ref{sec:experiments} presents and analyses the experimental results. Finally, Section~\ref{sec:conc} concludes the paper and discusses future work. 

\section{Background}
\label{sec:back}
This work intersects several research fields including GANs, synthetic data generation, optimisation, and reinforcement learning. We provide brief introductions to the key concepts and refer the reader to cited literature for further details. 

\subsection{Tabular GANs}
\label{sec:TGANs}
GANs~\cite{gan}, initially developed for image generation, are comprised of two neural network components: the \textit{generator} and the \textit{discriminator} that engage in a minimax game. The generator's objective is to synthesize images from noise, typically sampled from a uniform or Gaussian distribution, with the aim of deceiving the discriminator by producing the best approximation to the real image or data possible. Conversely, the discriminator endeavors to optimally classify images as real or synthetic. The objective functions for the discriminator and the generator are represented by the Equations~\ref{eq:loss-d} and \ref{eq:loss-g}, respectively.

\begin{equation} \label{eq:loss-d}
    \nabla_w[\frac{1}{n}\sum^n_{i=1} log(D_w(x^{(i)}))\cdot log(1-D_w(G(z_i)))]
\end{equation}

\begin{equation} \label{eq:loss-g}
    \nabla_{\theta}[\frac{1}{n}\sum^n_{i=1}log(D_{\color{black}w}(G(z_i)))],
\end{equation}
where $n$ is the batch size, $w$ and $\theta$ is the weight for the discriminator and the generator, respectively, $z$ is the noise vector sampled from the prior probability distribution, and $x^{(i)}$ is the sample from real data distribution. 

However, as previously noted, GANs were initially designed for image generation, which differs significantly from tabular data in terms of data structure and variable distributions. To accommodate these unique characteristics, the Conditional Tabular GAN (CTGAN)~\cite{ctgan} was developed. CTGAN introduces a conditional vector to the noise, representing the conditions of discrete variables, and employs a novel sampling strategy, \textit{training-by-sampling}, which comprehensively explores all values in categorical variables. For continuous variables with complex, multimodal distributions, CTGAN proposes a mode-specific normalisation as an alternative to min-max normalisation. A variational Gaussian mixture model (VGM) is utilized to estimate the number of modes for a continuous variable and the probability densities of these modes. When a value is sampled from a specific mode, it is also normalized according to that mode. In our proposed method, CTGAN serves as the foundation.

\subsection{Evolutionary GANs}
The innovative GAN framework enabled the generation of images from  probability distributions for the first time, subsequently garnering significant research interest and inspiring numerous enhancements, particularly with respect to loss function design. For example, least squares GANs (LSGANs)~\cite{lsgan} employ the least squares loss function in place of the original cross-entropy loss, resulting in improved image quality and training stability. Deep convolutional GANs (DCGAN)~\cite{dcgan} modify the loss function by maximising the log probability of the discriminator error, thereby mitigating the vanishing gradient problem. Gradient penalty~\cite{gp}, introduced in Wasserstein GAN~\cite{wgan}, is applied in our model as well, being added to the loss of the discriminator that enforces the discriminator's gradients to have a unit norm, thus effectively addressing the vanishing gradient issue. Evolutionary GANs (EGANs)~\cite{egan} integrate the loss functions of DCGAN, LSGAN, and original GANs by randomly selecting one of these three functions to train an offspring. EGAN serves as a component of the evolutionary process for SMOEGAN~\cite{smoegan}. 

\subsection{Multi-objective Optimisation}
\label{sec:nsga}
Multi-objective optimisation is concerned with the discovery of solutions to a problem that represent the ``best'' trade off between multiple conflicting objectives. More formally, a multi-objective problem can be defined as 
\begin{equation} \label{eq:mini}
\text{minimize} \hspace{0.5cm} f(\Vec{x}) \coloneqq [f_1(\Vec{x}),f_2(\Vec{x}),\dots,f_m(\Vec{x})],
\end{equation}
where $\Vec{x} = [x_1,x_2,\dots,x_n]$, $x_i\in\mathbb{R}$ represents the vector of decision variables (also known as solution or individual in the EA community), {\color{black}\textit{m} represents the number of} objectives, and $f_j:\mathbb{R}^n\rightarrow\mathbb{R}$ are the conflicting objective functions. 

There are a variety of optimisation algorithms to tackle a multi-objective problem~\cite{emmerich2018tutorial}. The algorithm employed in our study is based on the popular non-dominated sorting genetic algorithm II (NSGA-II)~\cite{nsga2}, a choice made by convenience. As the name implies, NSGA-II relies on the concept of non-dominated sorting (as opposed to decomposition or indicator-based) and crowding distance to drive the search for a well-spread set of trade off solutions. 

\vspace{+2mm}
\noindent\textbf{Non-dominated sorting} 
The notion of dominance drives this sorting approach. {\color{black}A solution $\Vec{x}$ is considered to dominate $\Vec{x'}$ (expressed as $\Vec{x} \prec \Vec{x'}$) if:

\begin{equation}
\begin{split}
\forall k \in {1, 2, \dots, m}, f_k(x) \leq f_k(x') \hspace{1 cm} \text{and} \\
\exists k \in {1, 2, \dots, m} \hspace{0.25 cm} \text{such that} \hspace{0.25 cm} f_k(x) < f_k(x')
\end{split}
\end{equation}
}
Given a population of solutions, non-dominated sorting iteratively identifies the set of non-dominated solutions and assigns them a rank of 1, then removes these solutions to identify the next set of non-dominated solutions assigned a rank of 2, and so on. This ranking can then be used to differentiate between the quality of solutions.  %proceeds as follows: For each individual $x_k, k=1,\dots,N$ in $P$, calculate the domination count $d_k$, representing the number of solutions that dominate $x_k$.%, and the set of solutions dominated by $x_i$, denoted as $S_i$. 

%Assign each individual a rank based on non-domination levels. Rank 1 corresponds to the non-dominated \textbf{front} $F_1$. A vector of decision variables $\Vec{x} \in X \subset \mathbb{R}^n$ is considered non-dominated with respect to $X$ if there does not exist another $\Vec{x}' \in X$ such that $f(\Vec{x}') \prec f(\Vec{x})$. Subsequently, remove $F_1$ from the population and repeat the process to determine $F_2, F_3$, etc., until all individuals are assigned a rank.

\vspace{+2mm}
\noindent\textbf{Crowding distance assignment}
The crowding distance allows NSGA-II to measure how close a solution is to its neighboring solutions in the objective space. This measure can then be used to differentiate between solutions of the same rank with solutions that are further away from their neighbors (i.e., located in a less crowded region of the objective space) being preferred. Formally, the crowding distance $l$ of a solution {\color{black}$\Vec{x}$ is computed as follows:
\begin{equation}\label{eq:crowd_dis}
l_{\Vec{x}} = \sum_{k=1}^m \frac{f_k(\Vec{y})-f_k(\Vec{z})}{f_{k,\max} - f_{k,\min}},
\end{equation}
where $f_{k,\max}$ and $f_{k,\min}$ represent the maximum and minimum values of the $k$-th objective function within the population, respectively, and $\Vec{y}$ and $\Vec{z}$ the two adjacent solutions of $\Vec{x}$ with respect to objective $m$ (forming a cuboid around $\Vec{x}$).} %The crowding distance serves as a metric for the density of solutions in the objective space. Individuals possessing lower ranks and higher crowding distances are chosen for the subsequent generation.

\section{Methodology}
\label{sec:method}
\begin{figure*}
    \centering
    \includegraphics[width=\textwidth]{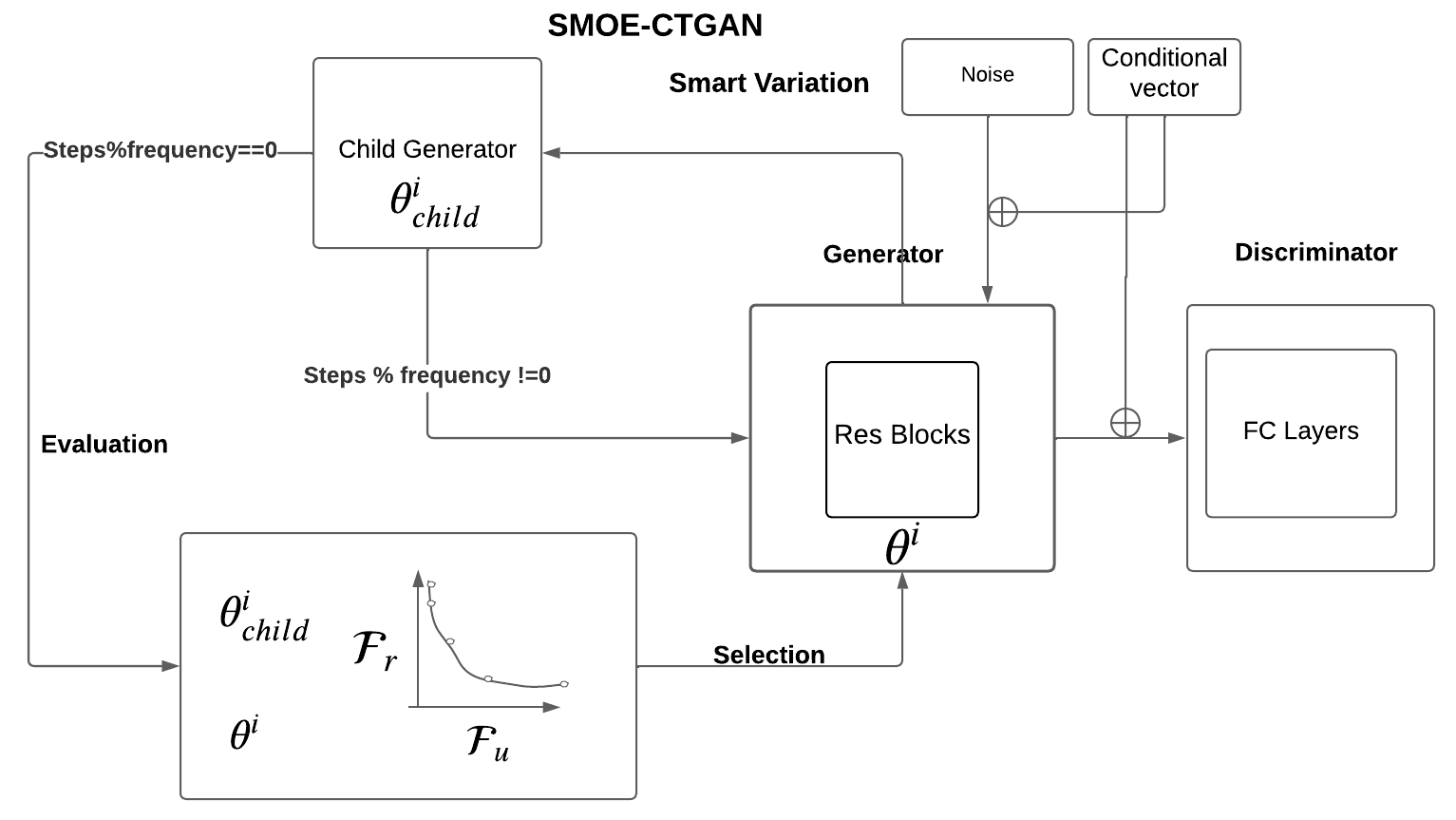}
    \caption[]{{\color{black}The training flow of SMOE-CTGAN. It begins with sampling noise from a normal distribution and creating a conditional vector from categorical data. Both are used by the generator to create data, which is then evaluated by a discriminator with the conditional vector included in its input. To get better and diverse offspring, three different loss functions and a deep reinforcement learning algorithm are used to select the best loss function to train the generator at each step given its evaluation values. Then a multi-objective selection operator is applied at regular intervals (select frequency) to choose the best candidates from a mix of parent and offspring solutions for the next generation, using non-dominated sorting and crowding distance during evaluation.}}
    \label{fig:paradigm}
\end{figure*}
Our SMOE-CTGAN comprises two primary components: CTGAN~\cite{ctgan} and the SMOE algorithm used by SMO-EGAN~\cite{smoegan}. SMO-EGAN was initially employed for image generation, so the metrics used in variation selection and multi-objective optimisation, as well as the general training paradigm, must be adapted and fine-tuned for generating tabular data. This adaptation is one of the major contributions of this work.  

In this section, we will provide a detailed overview of the model structure, loss and objective function design, training pipeline, and improvements for training speed, as illustrated in Figure~\ref{fig:paradigm}. Initially, noise is sampled from a normal distribution, and a conditional vector is derived on the basis of the distribution of the categorical data. Both are utilized by the generator to synthesize tabular data, which is subsequently classified by a discriminator. The aforementioned conditional vector, employed within the generator, is also incorporated into the discriminator's input. This is followed by an optimised variation process: the employment of three distinct loss functions in conjunction with a deep reinforcement learning algorithm. This combination is leveraged to select the optimal loss function to train the generator, with the aim of producing high-quality offspring. To mitigate time consumption and underfitting, a multi-objective selection operator is applied every $H$ steps, hereinafter referred to as the `select frequency'. During the evaluation phase, non-dominated sorting combined with crowding distance is employed to select the most favorable candidates for the succeeding generation from the pool of parent and offspring solutions.

Our objective is to demonstrate the effectiveness and enhancements achieved by applying the SMO-EGAN algorithm to CTGAN. Consequently, we maintain the SMOE-CTGAN backbone model structure identical to CTGAN, where both the discriminator and generator are neural networks comprising several fully-connected hidden layers. Furthermore, the settings for the conditional vector introduced, the Gumbel softmax~\cite{gumbel} activation for estimating gradients from categorical variables, and ten samples in each PAC~\cite{pacgan} in the discriminator all remain unaltered.

Rather than utilising the WGAN~\cite{wgan} losses (employed in CTGAN), we opt for the original cross-entropy loss function (Equation~\ref{eq:loss-d}) proposed in GAN~\cite{gan} to train the discriminator in our algorithm to collaborate with the three losses $L_{\text{minimax}}$~\cite{gan}, $L_{\text{heuristic}}$~\cite{dcgan}, and $L_{\text{least-square}}$~\cite{lsgan} used in evolutionary GANs (EGAN)~\cite{egan}. Same as the evolution in EGAN, at each step, a generator offspring is produced by applying one of the three loss functions to train for a batch. Let $M=\{L_{\text{minimax}}, L_{\text{heuristic}}, L_{\text{least-square}}\}$, but note that our GANs are conditional; thus, the final loss function of the generators is given by Equation~\ref{eq:m}. The advantages of employing these three losses instead of a single $L_{\text{minimax}}$ loss are threefold:
\begin{itemize}
    \item $L_{least-square}$ and $L_{heuristic}$ both do not suffer from vanishing gradient issues, which is a problem for $L_{minimax}$.
    \item The combination helps prevent model collapses during training, 
    \item The combination assists the generator in evolving during the training process.
\end{itemize}

\begin{equation} \label{eq:lminimax}
    L_{minimax}=\frac{1}{2}\mathbb{E}_{z~q(z)}[log(1-D(G(z)))]  
\end{equation}
\begin{equation} \label{eq:lheu}
    L_{heuristic}=-\frac{1}{2}\mathbb{E}_{z~q(z)}[log(D(G(z)))]  
\end{equation}
\begin{equation} \label{eq:lls}
    L_{least-square}=\mathbb{E}_{z~q(z)}[log(D(G(z))-1)^2]  
\end{equation}
\begin{equation} \label{eq:m}
    L_G=M^a+L_{condition}
\end{equation}

\begin{algorithm}
\caption{SMOE-CTGAN}\label{smoectgan_alg}

\KwIn{batch size $n$, population size $\mu$, number of objective functions $N_m$, discriminator's update steps per iteration $N_d$, Adam hyperparameters $\alpha,\beta_1,\beta_2$}

\textbf{Initialization:} Q-Function's parameters $Q$, discriminator's parameters $w$, generators' parameters $\Theta^1$, ..., $\Theta^{\mu}$, $\oplus$ as vector concatenating operator \\
Initialisation of transition batch $B=\{\emptyset\}$ \\
\For{\textup{number of training iterations}}{
    \For{$k\leftarrow 1$ \KwTo $N_d$}{
        Sample a batch $\vv{x}$ from dataset $P_{data}$ \\
        Generate noise batch $\vv{z}$ from normal distribution along with conditional vector ${c_1}$ and ${c_2}$ \\
        $\vv{z}\leftarrow \vv{z} \oplus c_1$ \tcp*{concatenate conditional vector}
        $\vv{x}\leftarrow \vv{x} \oplus c_2$ \tcp*{concatenate conditional vector}
        $g_w\leftarrow \nabla_w\left[\frac{1}{n}\sum^n_{i=1}\log D_w(x_i) +\right.$\
        $\left.\frac{1}{n}\sum^{\mu}_{j=1}\sum^{n/\mu}_{i=1}\log(1-D_w(G_j(z_i\oplus c_1^i))) +\right.$\
        $\left. Gradient\_Penalty\right]$ \tcp*{calculate the loss for discriminator}

        Update $w$ with Adam \\
    }
    \For{$j\leftarrow 1$ \KwTo $\mu$}{
        Generate noise batch $\vv{z}$ from a normal distribution along with conditional vector $c_1$ and $c_2$ \\
        $\Theta^j_{child}\leftarrow$ Smart Variation $(D,\Theta^j,\vv{z},\alpha,\beta_1,\beta_2,c_1,c_2,Q,B,\epsilon=10\%)$ \\
        Evaluate the new generator $\Theta^j_{child}$ \\
    }
    $PQ\leftarrow\Theta^1,...\Theta^{\mu},\Theta^1_{child},...,\Theta^{\mu}_{child}$ \\
    $\mathcal{PF}\leftarrow$ non\_dominated\_sorting$(PQ)$ \tcp*{start multi-objective sorting, using NSGA-II}
    crowding\_distance$(\mathcal{PF})$ \\
    $\Theta^1,...\Theta^{\mu}\leftarrow$ sorting$(\mathcal{PF})$ \\
    Sample random minibatch $b$ from $B$ \\
    Perform the gradient descent step on $Q$, using the samples in $b$ \\
}
\end{algorithm}
In the training pipeline, at each step, we first train the discriminator for $N_d$ steps, followed by training the generators to obtain offspring through Smart Variation, as detailed in Section~\ref{sec:smart}. In contrast to the training pipeline of SMO-EGAN, we sample noise in conjunction with conditional vectors, which are sampled from the distribution of categorical variables, and employ the training-by-sampling strategy~\cite{ctgan}, which is more adept at training conditional tabular data than iterating through the dataset from beginning to end for each epoch. Consequently, we need to concatenate conditional vectors to both the synthetic and real inputs for the discriminator, following the same procedure as in CTGAN~\cite{ctgan}. The conditional loss is computed using the cross-entropy loss in conjunction with the generator loss, taking the generator's output as the prediction and the conditional vector $c_1$ as labels. The two objectives in NSGA-II (explained in Section~\ref{sec:nsga}) are replaced by the marginal \textit{Targeted Correct Attribution Probability} (TCAP) measure $\mathcal{F}_r$ and the utility function $\mathcal{F}_u$, which will be introduced in Section~\ref{sub:evaluation}.

\begin{algorithm}
\caption{Smart Variation}\label{smart_var}

\KwIn{Generator's parameters $\Theta^j$, Noise Batch $\vv{z}$,Q-Function $Q$, Transition Batch $B$, Exploitation probability $\sigma$, Adam hyperparameters $\alpha,\beta_1,\beta_2$, conditional vector $c_1,c_2$, vector concatenating operator $\oplus$, conditional loss function $CE$}

$s^j\leftarrow\mathcal{F}_{ru}(\Theta^j)$ \tcp*{calculate utility and risk as input state for deep reinforcement learning}

\uIf{$rand < \epsilon $}{ \tcp*{$\epsilon$ greedy strategy}
    $a\leftarrow$ random loss functions
}
\Else{
    $a\leftarrow \arg\max_a Q(s^j,a)$ \tcp*{choose the loss function to maximize reward}
}

$fake\leftarrow G_{\theta^j}(\vv{z}\oplus c_1)$ \tcp*{generate output from generator}

$grad\leftarrow \nabla [M^a(fake,c_1,c_2)+CE(fake,c_1)]$ \tcp*{calculate loss of generator using chosen loss function and add a condition loss}

$\Theta^j_{child}\leftarrow Adam(grad,\Theta^j,\alpha,\beta_1,\beta_2)$

$r\leftarrow R(\Theta^j,\Theta^j_{child})$ \tcp*{reward function}

$s^j_{child}=\mathcal{F}_{ru}(\Theta^j_{child})$ \tcp*{update state values by calculating utility and risk of the child generated}

$a_{next}\leftarrow \arg\max_a Q(s^j_{child},a)$

Store transition $(s^j,a,r,s^j_{child},a_{next})$ in $B$ \tcp*{Store the transitions for neural network training}

\Return $\Theta^j_{child}$
\end{algorithm}

\subsection{Smart Variation}
\label{sec:smart}
\begin{figure}
    \centering
    \includegraphics[width=0.5\textwidth]{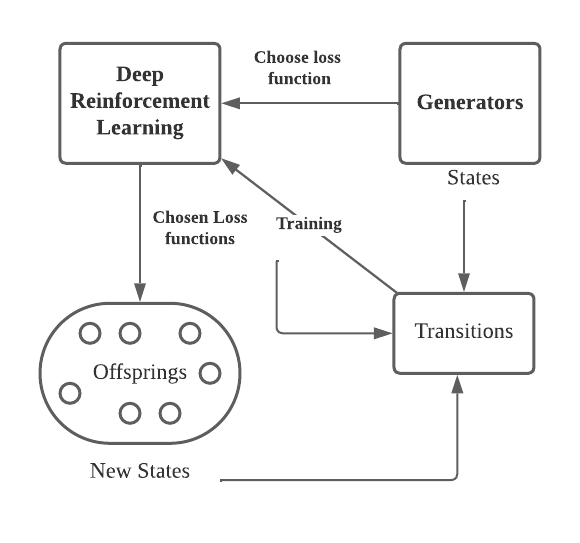}
    \caption{Smart Variation: The loss functions used to train generators are selected by a deep reinforcement learning module, the choices of loss functions are actions. The states of generators are encoded by utility and risk values. With the values of states, new states and actions are stored in transitions and then passed to deep reinforcement learning for its training.}
    \label{fig:smart_var}
\end{figure}

Algorithm~\ref{smart_var} presents the pseudocode for Smart Variation to generate new generators (offspring). This component is primarily implemented using $Q$-Learning~\cite{sutton2018reinforcement}, a model-free reinforcement learning algorithm. A multi-layer, fully-connected neural network (NN) serves as the agent, taking the state as input and predicting appropriate actions to maximize the reward. The state is encoded as $[Risk, Utility] = [\mathcal{F}_r, \mathcal{F}_u]$, with both variables already in the range of 0 to 1, which is advantageous for feeding into the NN. Additionally, we employ an $\epsilon$-greedy strategy to maintain exploration in the target space. We follow the approach of SMO-EGAN~\cite{smoegan} and use one objective (instead of both) in the reward function of the reinforcement learning algorithm (future work can look at a multi-reward setup): we decided to base the reward function on the utility objective because of the preference to create datasets with a higher utility (as opposed to lower disclosure risk). More formally, our reward function is defined as
\begin{equation}
    R(\Theta^j,\Theta^j_{\text{child}}) = \begin{cases}
         1 & \text{if } \mathcal{F}_u(\Theta^j_{\text{child}}) > \mathcal{F}_u(\Theta^j) \\
         0 & \text{otherwise.}
    \end{cases}
\end{equation}
In plain English, we assign a reward of 1 if the new state is associated with a higher utility. After accumulating enough of the transition, it will be stacked as a batch to train the NN. The whole process is also illustrated in Figure~\ref{fig:smart_var}. 

\subsection{Selection}
The selection process for each step relies on non-dominated sorting and crowding distance.%, aiming to select individuals with superior and distinct performance according to Pareto dominance. Non-dominated Sorting ranks performance based on multiple objectives, while Crowding Distance aims to create a less dense population. 
At the start of training, instances with low risk but impractically low utility (failures) may arise due to underfitting because an underfitted model can have large negative risk, impacting the accuracy of selection. To mitigate this issue, we adopt a strategy of clipping the risk value to a minimum threshold, based on training behavior of each dataset.

\section{Experiments}
\label{sec:experiments}
This section describes the experimental settings and considerations, followed by an analysis of the experimental results.
\subsection{Evaluation}
\label{sub:evaluation}
To calculate the overall utility, we consider the average Confidence Interval Overlap (CIO)~\cite{cio} and the Ratio of Counts/Estimates~(ROC) as outlined in the previous work~\cite{framework}. We employ 95\% confidence intervals to compute the CIO, selecting the coefficients from regression models built on both the original and synthetic datasets. The ROC represents the ratio of categorical variable occurrences in the original and synthetic data, utilising frequency tables and cross-tabulations. These metrics are formulated as follows:
\begin{equation} \label{eq:cio}
    CIO=\frac{1}{2} \left\{ \frac{min(u_o,u_s)-max(l_o,l_s)}{u_o-l_o} + \frac{min(u_o,u_s)-max(l_o,l_s)}{u_s-l_s} \right\}
\end{equation}
\begin{equation}
    ROC = \frac{min(y_{orig},y_{synth})}{max(y_{orig},y_{synth})}
\end{equation}
For CIO, we compute the average across all coefficients of the regression, using $u_o, l_o$ and $u_s, l_s$ to represent the upper and lower bounds of the confidence intervals for the original and synthetic tabular data, respectively. The CIO values range from negative values (indicating no overlap) to 1 (complete overlap), higher values signify better utility. To ensure a fair comparison, we use the same variable sets as in~\cite{framework}, with detailed variables available in their work's appendix.

For ROC, we compute the average of univariate and all possible combinations in bivariate for each categorical feature chosen. Disclosure risk is assessed using the Targeted Correct Attribution Probability (TCAP)~\cite{tcap}, which estimates the likelihood of an intruder correctly inferring sensitive attributes about an individual, given the knowledge that the individual is in the dataset and information about a specified set of attributes. With a value range of 0 to 1, a higher TCAP indicates a higher risk. We normalize the TCAP following the procedure in~\cite{framework}, resulting in a new range from the risk of random guess (0) to 1, allowing for negative TCAP values, which simulates an adversary doing worse than randomly drawing from the distribution of the target. We use the same subset of variables as in the Appendix of~\cite{framework} for TCAP.

To compute TCAP and utility function, we generate a synthetic dataset with the same size as the original dataset for evaluation in each epoch. These metrics are also utilized as objective functions in our multi-objective optimisation process. However, computing these metrics for large datasets can be extremely time-consuming, so we use only 500 samples from both synthetic and relevant original datasets for calculations during the multi-objective optimisation step. Despite the time-saving approach, an epoch's training time may still be large. As a result, we further reduce the frequency of multi-objective optimisation to once every $H$ steps ($H$ often being 4, 8, or 16) instead of once per step, as in the original SMO-EGAN, depending on the dataset size. These adjustments lead to acceptable training times and maintain superior performance while stabilising the training process, as small fluctuations in evaluation metrics for each step might otherwise misguide the selection process.

\subsection{Select Frequency}
In this study, we implement a multi-objective optimisation procedure by decreasing the selection frequency, which entails executing the optimisation and selection processes for every $H$ training steps. An experiment was conducted to examine the implications of reduced frequency on the performance of the model. The result of this experiment is shown in Table~\ref{tab:select}. It can be seen that as the value of $H$ increases, the utility value does not exhibit a significant decrease, and the risk remains relatively constant. This approach enables a considerable reduction in training time without compromising the quality of the results. Furthermore, it has been ascertained that employing a smaller value of $H$ is more conducive for larger datasets, whereas a larger value of $H$ is optimal for datasets comprising fewer than 50,000 cases. These findings have been incorporated into our final results, thus enhancing the overall efficiency of the model.
\begin{table}[h]
\centering
\begin{tabular}{ccc}
\hline
Select frequency $H$ & Utility score & TCAP \\ \hline
2  & 0.545  & 0.048 \\ 
4  & 0.533  & -0.084  \\ 
8  & 0.547  & 0.027  \\ 
16  & 0.528  & 0.036  \\ \hline
\end{tabular}
\caption{Results of SMO-EGAN for different select frequencies on the Canada census dataset.}
\label{tab:select}
\end{table}

\subsection{Results and Analysis}
\begin{table*}[ht]
    \centering
    \setlength{\tabcolsep}{8pt} % Adjust column spacing
    \begin{tabularx}{\textwidth}{p{1.7cm}p{5.3cm}p{1.7cm}p{1.8cm}p{1.6cm}p{1.9cm}}
        \toprule
        Dataset & Synthesizer & Overall Utility & Risk (marginal TCAP) & Population size & MO frequency (steps/MO) $H$ \\
        \midrule
        \multirow{5}{*}{UK 1991} & CTGAN~\cite{framework} & \textbf{0.514} & 0.371 & & \\
                                 & Synthpop~\cite{framework} & 0.774 & 0.516 & & \\
                                 & DataSynthesizer (No DP)~\cite{framework} & 0.643 & 0.440 & & \\
                                 & CTGAN with Improvement Score & 0.482 & 0.041 & & \\
                                 & SMOE-CTGAN with Improvement Score & \textbf{0.493} & \textbf{0.018} & 4 & 8 \\
                                 & SMOE-CTGAN with one objective & \textbf{0.499} & 0.067 & 4 & 8 \\
                                 & SMOE-CTGAN & \textbf{0.493} & 0.094 & 4 & 8 \\
                                 & SMOE-CTGAN & 0.487 & 0.050 & 8 & 8 \\
        \midrule
        \multirow{5}{*}{Canada 2011} & CTGAN~\cite{framework} & 0.495 & 0.165 & & \\
                                 & Synthpop~\cite{framework} & 0.830 & 0.294 & & \\
                                 & DataSynthesizer (No DP)~\cite{framework} & 0.688 & 0.231 & & \\
                                 & CTGAN with Improvement Score & 0.499 & \textbf{-0.041} & & \\
                                 & SMOE-CTGAN with Improvement Score & 0.516 & \textbf{-0.005} & 4 & 16 \\
                                 & SMOE-CTGAN with one objective & 0.508 & 0.116 & 4 & 16 \\
                                 & SMOE-CTGAN & 0.528 & 0.036 & 4 & 16 \\
                                 & SMOE-CTGAN & \textbf{0.553} & 0.044 & 8 & 16 \\
        \midrule
        \multirow{5}{*}{Fiji 2007} & CTGAN~\cite{framework} & 0.469 & 0.439 & & \\
                                 & Synthpop~\cite{framework} & 0.816 & 0.555 & & \\
                                 & DataSynthesizer (No DP)~\cite{framework} & 0.727 & 0.526 & & \\
                                 & CTGAN with Improvement Score & 0.414 & \textbf{-0.011} & & \\
                                 & SMOE-CTGAN with Improvement Score & 0.437 & \textbf{0.114} & 4 & 8 \\
                                 & SMOE-CTGAN with one objective & \textbf{0.479} & 0.389 & 4 & 8 \\
                                 & SMOE-CTGAN & \textbf{0.475} & 0.320 & 4 & 8 \\
                                 & SMOE-CTGAN & 0.457 & 0.172 & 8 & 8 \\
                                 
        \midrule
        \multirow{5}{*}{Rwanda 2012} & CTGAN~\cite{framework} & 0.430 & 0.412 & & \\
                                 & Synthpop~\cite{framework} & 0.752 & 0.437 & & \\
                                 & DataSynthesizer (No DP)~\cite{framework} & 0.720 & 0.413 & & \\
                                 & CTGAN with Improvement Score & 0.434 & -0.007 & & \\
                                 & SMOE-CTGAN with Improvement Score & 0.449 & \textbf{-0.033} & 4 & 16 \\
                                 & SMOE-CTGAN with one objective & 0.470 & 0.391 & 4 & 16 \\
                                 & SMOE-CTGAN & 0.477 & 0.321 & 4 & 16 \\
                                 & SMOE-CTGAN & \textbf{0.505} & 0.172 & 8 & 16 \\
        \bottomrule
    \end{tabularx}
    \caption{Results comparison between our CTGAN with Improvement Score, SMOE-CTGAN, and results from~\cite{framework}. {\color{black}Note that the difference between the two SMOE-CTGAN (bottom two row for each dataset) is the population size used (4 vs 8). Values highlighted in bold indicate similarly good results amongst deep-learning based methods. }}
    \label{tab:performance}
\end{table*}

We trained the model using the original CTGAN as our baseline. In line with~\cite{framework}, we evaluate our method on four census datasets, including those from the UK~\cite{uk_data}, Canada~\cite{ipums}, Rwanda~\cite{ipums}, and Fiji~\cite{ipums}. For each dataset, we train five CTGANs with different seeds, and the average performance is considered as the overall performance. That is, if not otherwise stated, any results shown are average results across five runs. We maintain the default hyperparameters in CTGAN, which include the Adam optimiser with a learning rate of 0.0002, weight decay of 1e-6, $\beta_1$ and $\beta_2$ of 0.5 and 0.9, respectively, and a batch size of 500 for both discriminator and generator. These settings are also applied to the optimisers in our SMOE-CTGAN, replacing the settings from SMO-EGAN to better suit tabular data.

During the training process, we observed that in the early epochs, all models achieve relatively high utility while simultaneously exhibiting markedly lower risk compared to the scores presented in~\cite{framework}.
\begin{figure}
    \centering
    \includegraphics[width=0.45\textwidth]{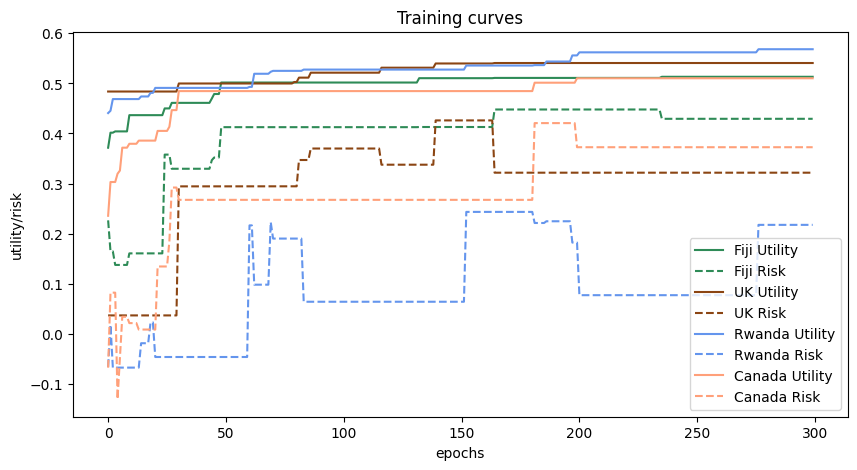}
    \caption{Training curves on all census datasets by the proposed approach, SMOE-CTGAN.}
    \label{fig:training_curves}
\end{figure}
As the training progresses, the optimal utility demonstrates only a marginal improvement, while the risk increases rapidly by a significant margin. During the training of SMOE-CTGAN, we observed similar phenomena, as the example trainings illustrated in Figure~\ref{fig:training_curves}. This figure presents the initial scores, the best scores after 60 epochs and 300 epochs, respectively, and the trade-off between them as the number of training epochs increases.

This raises the question of whether this trade-off is worthwhile. To address this, we introduce a simple Improvement Score to determine if the model at the current epoch outperforms the current best model. This is expressed in the following equation:
\begin{equation}\label{eq:is}
    Improvement\_score = \lambda \cdot(\mathcal{F}_u^i - \mathcal{F}_u^{best}) + max(\mathcal{F}_r^{best},0) - max(\mathcal{F}_r^i,0), 
\end{equation}
where $\mathcal{F}_u$ and $\mathcal{F}_r$ denote the utility and risk functions, respectively, and $\lambda$ is a weighting factor that adjusts the trade-off between an increase in utility and a decrease in risk, with $i$ representing the $i^{th}$ epoch. If the Improvement Score is greater than 0, then the model at the current $i^{th}$ epoch outperforms the best model obtained thus far. For example, if $\lambda = 2$, an increase of 0.2 in utility can result in a 0.4 increase in risk (twice the magnitude). Incorporating this Improvement Score, we achieved numerous promising results, as shown in Table~\ref{tab:performance}, where MO denotes the multi-objective optimisation operator. We observed that the fluctuations in risk values below zero are excessively intense, particularly during the early stages of training, which is the result of underfitting and random data synthesis. Consequently, we clip the risk value to 0 if it has a negative value. This also makes substantive sense. The choice of setting zero is based on an assumption about the information that the adversary already possesses. Settings below this value equate to the adversary learning less than they already know which clearly give them no advantage!

As is evident from Table~\ref{tab:performance}, our model with the Improvement Score consistently achieves low risk levels, approaching zero or even negative values. For smaller datasets, such as the Canada and Rwanda Ccnsus, which contain approximately 30,000 samples, SMOE-CTGAN achieves a higher utility compared to the CTGAN presented in~\cite{framework}, while maintaining considerably low risk due to the use of the Improvement Score. In scenarios where the Improvement Score is not utilized, the utility score of SMOE-CTGAN further increases; however, this comes at the expense of rapidly escalating risk, with the exception of the UK and Canada census datasets. Despite this increase, the risk remains substantially lower than that of CTGAN (see~\cite{framework}) demonstrating the efficacy of multi-objective optimisation for synthesising tabular data. In summary, SMOE-CTGAN achieves better overall utility compared to the CTGAN results reported in~\cite{framework}, and offers a distinct advantage in risk reduction while simultaneously achieving higher utility scores. With the aid of the Improvement Score, our approach achieves exceedingly low risk levels around zero and consistently outperforms the CTGAN with the Improvement Score. 
{\color{black}Furthermore, we ran an ablation study of using only one objective (utility) for our SMOE-CTGAN. The results indicate that without the multi-objective optimization step, our models have largely increased risk, as well as achieved lower utility compared to our best model.}

\subsection{Pareto Front with Improvement Score Analysis}
\begin{figure}[t!]
    \centering
    \includegraphics[width=0.5\textwidth]{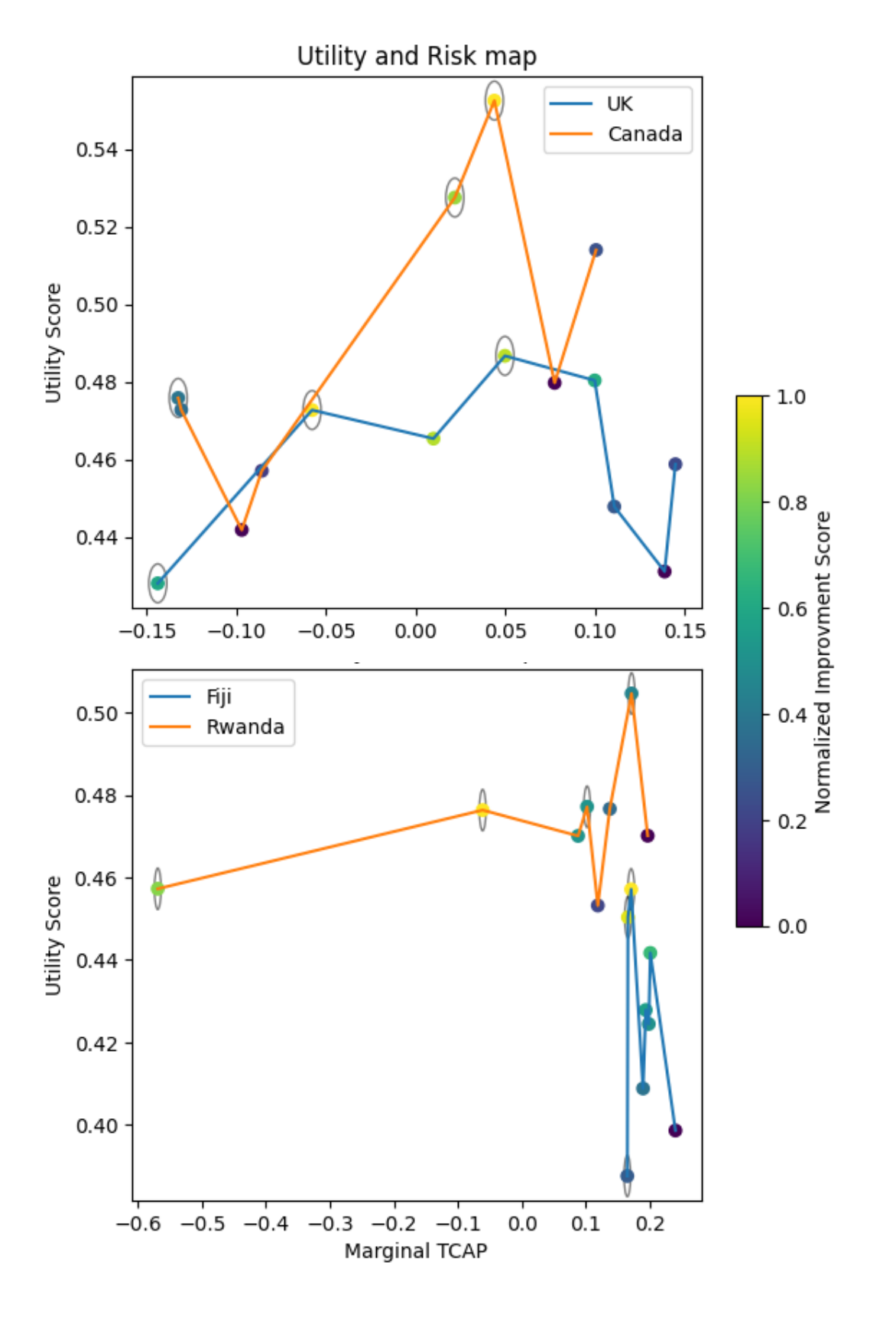}
    \caption{Final population (of 8 solutions) of solutions from a single run plotted against the two objectives and the normalized Improvement Score. The solutions surrounded by circles are the non-dominated solutions.}
    \label{fig:urmap}
\end{figure}

Figure~\ref{fig:urmap} shows the population of SMOE-CTGAN after training on all datasets. We can use the Improvement Score to help select a final solution (generator) from the population. The yellow points represent the generators ultimately chosen if the Improvement Score is applied, and the points with a circle are non-dominated points. {\color{black}We also show the dominated solutions to provide an idea of the spread of the final population and also show the the Improvement Score is lowest for dominated solutions, which is a desired effect.}

It is evident that the yellow points are predominantly associated with the highest utility. The points with highest utility is selected as the best if the Improvement Score is not applied. This demonstrates the effectiveness of implementing the Improvement Score for generator selection with competitively high utility and lower risk. Moreover, the $\lambda$ value in the Improvement Score can be adjusted according to the desired emphasis on utility relative to risk levels; in our case, we use a value of 2.

\section{Conclusion}
\label{sec:conc}
In this study, we first analyzed the training process of GANs for synthesising tabular data and observed that, in the early stages of training, GANs can achieve competitive  utility of a synthetic dataset with significantly lower risk of re-identification. However, as training progresses, the model pays a considerable price in terms of increased risk for only marginal improvements in utility. Consequently, we introduced a simple metric called Improvement Score to evaluate whether a model outperforms its predecessor by comparing improvements in both utility and risk, resulting in models with substantially reduced risk.

Subsequently, we proposed a novel GAN framework for synthesising tabular data called SMOE-CTGAN, which applies multi-objective optimisation to CTGAN based on utility and risk functions specific to tabular data. We further enhanced the training and evaluation paradigm to significantly reduce training time without compromising performance. SMOE-CTGAN outperforms CTGAN with Improvement Score on multiple census datasets from different countries in terms of both lower risk and higher utility. Moreover, in the absence of the Improvement Score, the models attained an even higher utility, albeit at the cost of a rapid increase in risk. Nonetheless, the risk levels remained lower than those reported in previous work~\cite{framework}. Although the utility of our models do not surpass that of Synthpop and DataSynthesizer, our approach demonstrates competitive utility with near-zero marginal TCAP, indicating the promising potential of employing GANs or neural networks for generating tabular data with high utility and negligible marginal TCAP, akin to random guesses. This gives GANs an exceptional advantage compared to other data synthesis methods. Furthermore, it is essential to recognize the phenomenon wherein the model achieves a substantial utility score accompanied by exceptionally low risk during the initial stages of training. This observation may aid in the future design of generative neural networks, allowing them to better highlight their advantages and strengths in comparison to alternative methods.

The application of multi-objective optimsation to GANs is relatively new. Future research can explore in more detail the sensitivity of the different algorithm parameters on performance. Improved performance can perhaps also be achieved by applying the proposed Improvement Score at each selection step of the multi-objective optimiser. %It could also be worthwhile refining the metrics used in tabular data, such as using a multi-reward function within the deep reinforcement learning algorithm. 

\begin{comment}
1. test multi-objective in the reward function
2. can be downloaded from github: xxx. 
3. move the caption of figure one to methodology section
4. cite SMOEGAN in my reward function as it uses only one objective, same as SMOEGAN
5. change the font size in figure 1 to make it more consistent. 
6. update figure 4, using population 8, consistent values and positive y-axis
\end{comment}
%\section{Appendices}

\bibliographystyle{ACM-Reference-Format}
\bibliography{Main}


\begin{thebibliography}{38}

%%% the Web address.

\ifx \showCODEN    \undefined \def \showCODEN     #1{\unskip}     \fi
\ifx \showDOI      \undefined \def \showDOI       #1{#1}\fi
\ifx \showISBNx    \undefined \def \showISBNx     #1{\unskip}     \fi
\ifx \showISBNxiii \undefined \def \showISBNxiii  #1{\unskip}     \fi
\ifx \showISSN     \undefined \def \showISSN      #1{\unskip}     \fi
\ifx \showLCCN     \undefined \def \showLCCN      #1{\unskip}     \fi
\ifx \shownote     \undefined \def \shownote      #1{#1}          \fi
\ifx \showarticletitle \undefined \def \showarticletitle #1{#1}   \fi
\ifx \showURL      \undefined \def \showURL       {\relax}        \fi
% invisible to TeX
\providecommand\bibfield[2]{#2}
\providecommand\bibinfo[2]{#2}
\providecommand\natexlab[1]{#1}
\providecommand\showeprint[2][]{arXiv:#2}

\bibitem[\protect\citeauthoryear{Albuquerque, Monteiro, Doan, Considine, Falk,
  and Mitliagkas}{Albuquerque et~al\mbox{.}}{2019}]%
        {albuquerque2019multiobjective}
\bibfield{author}{\bibinfo{person}{Isabela Albuquerque}, \bibinfo{person}{Joao
  Monteiro}, \bibinfo{person}{Thang Doan}, \bibinfo{person}{Breandan
  Considine}, \bibinfo{person}{Tiago Falk}, {and} \bibinfo{person}{Ioannis
  Mitliagkas}.} \bibinfo{year}{2019}\natexlab{}.
\newblock \showarticletitle{Multi-objective training of Generative Adversarial
  Networks with multiple discriminators}. In
  \bibinfo{booktitle}{\emph{Proceedings of the 36th International Conference on
  Machine Learning}} \emph{(\bibinfo{series}{Proceedings of Machine Learning
  Research}, Vol.~\bibinfo{volume}{97})},
  \bibfield{editor}{\bibinfo{person}{Kamalika Chaudhuri} {and}
  \bibinfo{person}{Ruslan Salakhutdinov}} (Eds.). \bibinfo{publisher}{PMLR},
  \bibinfo{pages}{202--211}.
\newblock
\urldef\tempurl%
\url{https://proceedings.mlr.press/v97/albuquerque19a.html}
\showURL{%
\tempurl}


\bibitem[\protect\citeauthoryear{Arjovsky, Chintala, and Bottou}{Arjovsky
  et~al\mbox{.}}{2017}]%
        {wgan}
\bibfield{author}{\bibinfo{person}{Martin Arjovsky}, \bibinfo{person}{Soumith
  Chintala}, {and} \bibinfo{person}{Léon Bottou}.}
  \bibinfo{year}{2017}\natexlab{}.
\newblock \bibinfo{title}{Wasserstein GAN}.
\newblock
\newblock
\urldef\tempurl%
\url{https://doi.org/10.48550/ARXIV.1701.07875}
\showDOI{\tempurl}


\bibitem[\protect\citeauthoryear{Arora, Ge, Liang, Ma, and Zhang}{Arora
  et~al\mbox{.}}{2017}]%
        {arora2017generalization}
\bibfield{author}{\bibinfo{person}{Sanjeev Arora}, \bibinfo{person}{Rong Ge},
  \bibinfo{person}{Yingyu Liang}, \bibinfo{person}{Tengyu Ma}, {and}
  \bibinfo{person}{Yi Zhang}.} \bibinfo{year}{2017}\natexlab{}.
\newblock \showarticletitle{Generalization and equilibrium in generative
  adversarial nets (gans)}. In \bibinfo{booktitle}{\emph{International
  conference on machine learning}}. PMLR, \bibinfo{pages}{224--232}.
\newblock


\bibitem[\protect\citeauthoryear{Baioletti, Bari, Poggioni, and
  Coello}{Baioletti et~al\mbox{.}}{2021}]%
        {smoegan}
\bibfield{author}{\bibinfo{person}{Marco Baioletti},
  \bibinfo{person}{Gabriele~Di Bari}, \bibinfo{person}{Valentina Poggioni},
  {and} \bibinfo{person}{Carlos Artemio~Coello Coello}.}
  \bibinfo{year}{2021}\natexlab{}.
\newblock \showarticletitle{Smart Multi-Objective Evolutionary GAN}. In
  \bibinfo{booktitle}{\emph{2021 IEEE Congress on Evolutionary Computation
  (CEC)}}. \bibinfo{pages}{2218--2225}.
\newblock
\urldef\tempurl%
\url{https://doi.org/10.1109/CEC45853.2021.9504858}
\showDOI{\tempurl}


\bibitem[\protect\citeauthoryear{Center}{Center}{2020}]%
        {ipums}
\bibfield{author}{\bibinfo{person}{Minnesota~Population Center}.}
  \bibinfo{year}{2020}\natexlab{}.
\newblock \showarticletitle{Integrated Public Use Microdata Series,
  International: Version 7.3 [dataset].}
\newblock \bibinfo{journal}{\emph{Minneapolis, MN: IPUMS. IPUMs Census Data}}
  (\bibinfo{year}{2020}).
\newblock


\bibitem[\protect\citeauthoryear{Deb, Pratap, Agarwal, and Meyarivan}{Deb
  et~al\mbox{.}}{2002}]%
        {nsga2}
\bibfield{author}{\bibinfo{person}{Kalyanmoy Deb}, \bibinfo{person}{Amrit
  Pratap}, \bibinfo{person}{Sameer Agarwal}, {and} \bibinfo{person}{T.
  Meyarivan}.} \bibinfo{year}{2002}\natexlab{}.
\newblock \showarticletitle{A fast and elitist multiobjective genetic
  algorithm: NSGA-II}.
\newblock \bibinfo{journal}{\emph{IEEE Transactions on Evolutionary
  Computation}} \bibinfo{volume}{6}, \bibinfo{number}{2}
  (\bibinfo{year}{2002}), \bibinfo{pages}{182--197}.
\newblock
\urldef\tempurl%
\url{https://doi.org/10.1109/4235.996017}
\showDOI{\tempurl}


\bibitem[\protect\citeauthoryear{Drechsler and Reiter}{Drechsler and
  Reiter}{2010}]%
        {intruders}
\bibfield{author}{\bibinfo{person}{Jörg Drechsler} {and}
  \bibinfo{person}{Jerome~P. Reiter}.} \bibinfo{year}{2010}\natexlab{}.
\newblock \showarticletitle{Sampling With Synthesis: A New Approach for
  Releasing Public Use Census Microdata}.
\newblock \bibinfo{journal}{\emph{J. Amer. Statist. Assoc.}}
  \bibinfo{volume}{105}, \bibinfo{number}{492} (\bibinfo{year}{2010}),
  \bibinfo{pages}{1347--1357}.
\newblock
\urldef\tempurl%
\url{https://doi.org/10.1198/jasa.2010.ap09480}
\showDOI{\tempurl}
\showeprint{https://doi.org/10.1198/jasa.2010.ap09480}


\bibitem[\protect\citeauthoryear{Drefs, Guiraud, and Lücke}{Drefs
  et~al\mbox{.}}{2022}]%
        {drefs2022evolutionary}
\bibfield{author}{\bibinfo{person}{Jakob Drefs}, \bibinfo{person}{Enrico
  Guiraud}, {and} \bibinfo{person}{Jörg Lücke}.}
  \bibinfo{year}{2022}\natexlab{}.
\newblock \showarticletitle{Evolutionary Variational Optimization of Generative
  Models}.
\newblock \bibinfo{journal}{\emph{Journal of Machine Learning Research}}
  \bibinfo{volume}{23}, \bibinfo{number}{21} (\bibinfo{year}{2022}),
  \bibinfo{pages}{1--51}.
\newblock
\urldef\tempurl%
\url{http://jmlr.org/papers/v23/20-233.html}
\showURL{%
\tempurl}


\bibitem[\protect\citeauthoryear{Emmerich and Deutz}{Emmerich and
  Deutz}{2018}]%
        {emmerich2018tutorial}
\bibfield{author}{\bibinfo{person}{Michael~TM Emmerich} {and}
  \bibinfo{person}{Andr{\'e}~H Deutz}.} \bibinfo{year}{2018}\natexlab{}.
\newblock \showarticletitle{A tutorial on multiobjective optimization:
  fundamentals and evolutionary methods}.
\newblock \bibinfo{journal}{\emph{Natural computing}}  \bibinfo{volume}{17}
  (\bibinfo{year}{2018}), \bibinfo{pages}{585--609}.
\newblock


\bibitem[\protect\citeauthoryear{Goodfellow, Pouget-Abadie, Mirza, Xu,
  Warde-Farley, Ozair, Courville, and Bengio}{Goodfellow et~al\mbox{.}}{2014}]%
        {gan}
\bibfield{author}{\bibinfo{person}{Ian Goodfellow}, \bibinfo{person}{Jean
  Pouget-Abadie}, \bibinfo{person}{Mehdi Mirza}, \bibinfo{person}{Bing Xu},
  \bibinfo{person}{David Warde-Farley}, \bibinfo{person}{Sherjil Ozair},
  \bibinfo{person}{Aaron Courville}, {and} \bibinfo{person}{Yoshua Bengio}.}
  \bibinfo{year}{2014}\natexlab{}.
\newblock \showarticletitle{Generative Adversarial Nets}. In
  \bibinfo{booktitle}{\emph{Advances in Neural Information Processing
  Systems}}, \bibfield{editor}{\bibinfo{person}{Z.~Ghahramani},
  \bibinfo{person}{M.~Welling}, \bibinfo{person}{C.~Cortes},
  \bibinfo{person}{N.~Lawrence}, {and} \bibinfo{person}{K.Q. Weinberger}}
  (Eds.), Vol.~\bibinfo{volume}{27}. \bibinfo{publisher}{Curran Associates,
  Inc.}
\newblock
\urldef\tempurl%
\url{https://proceedings.neurips.cc/paper/2014/file/5ca3e9b122f61f8f06494c97b1afccf3-Paper.pdf}
\showURL{%
\tempurl}


\bibitem[\protect\citeauthoryear{Grantham, Mukaidaisi, Ooi, Ghaemi, Tchagang,
  and Li}{Grantham et~al\mbox{.}}{2022}]%
        {grantham2022deep}
\bibfield{author}{\bibinfo{person}{Karl Grantham}, \bibinfo{person}{Muhetaer
  Mukaidaisi}, \bibinfo{person}{Hsu~Kiang Ooi},
  \bibinfo{person}{Mohammad~Sajjad Ghaemi}, \bibinfo{person}{Alain Tchagang},
  {and} \bibinfo{person}{Yifeng Li}.} \bibinfo{year}{2022}\natexlab{}.
\newblock \showarticletitle{Deep Evolutionary Learning for Molecular Design}.
\newblock \bibinfo{journal}{\emph{IEEE Computational Intelligence Magazine}}
  \bibinfo{volume}{17}, \bibinfo{number}{2} (\bibinfo{year}{2022}),
  \bibinfo{pages}{14--28}.
\newblock
\urldef\tempurl%
\url{https://doi.org/10.1109/MCI.2022.3155308}
\showDOI{\tempurl}


\bibitem[\protect\citeauthoryear{Gulrajani, Ahmed, Arjovsky, Dumoulin, and
  Courville}{Gulrajani et~al\mbox{.}}{2017}]%
        {gp}
\bibfield{author}{\bibinfo{person}{Ishaan Gulrajani}, \bibinfo{person}{Faruk
  Ahmed}, \bibinfo{person}{Martin Arjovsky}, \bibinfo{person}{Vincent
  Dumoulin}, {and} \bibinfo{person}{Aaron~C Courville}.}
  \bibinfo{year}{2017}\natexlab{}.
\newblock \showarticletitle{Improved Training of Wasserstein GANs}. In
  \bibinfo{booktitle}{\emph{Advances in Neural Information Processing
  Systems}}, \bibfield{editor}{\bibinfo{person}{I.~Guyon},
  \bibinfo{person}{U.~Von Luxburg}, \bibinfo{person}{S.~Bengio},
  \bibinfo{person}{H.~Wallach}, \bibinfo{person}{R.~Fergus},
  \bibinfo{person}{S.~Vishwanathan}, {and} \bibinfo{person}{R.~Garnett}}
  (Eds.), Vol.~\bibinfo{volume}{30}. \bibinfo{publisher}{Curran Associates,
  Inc.}
\newblock
\urldef\tempurl%
\url{https://proceedings.neurips.cc/paper/2017/file/892c3b1c6dccd52936e27cbd0ff683d6-Paper.pdf}
\showURL{%
\tempurl}


\bibitem[\protect\citeauthoryear{Jang, Gu, and Poole}{Jang
  et~al\mbox{.}}{2017}]%
        {gumbel}
\bibfield{author}{\bibinfo{person}{Eric Jang}, \bibinfo{person}{Shixiang Gu},
  {and} \bibinfo{person}{Ben Poole}.} \bibinfo{year}{2017}\natexlab{}.
\newblock \showarticletitle{Categorical Reparameterization with
  Gumbel-Softmax}. In \bibinfo{booktitle}{\emph{International Conference on
  Learning Representations}}.
\newblock
\urldef\tempurl%
\url{https://openreview.net/forum?id=rkE3y85ee}
\showURL{%
\tempurl}


\bibitem[\protect\citeauthoryear{Jennifer and Mark}{Jennifer and Mark}{2019}]%
        {tcap}
\bibfield{author}{\bibinfo{person}{Taub Jennifer} {and} \bibinfo{person}{Elliot
  Mark}.} \bibinfo{year}{2019}\natexlab{}.
\newblock \showarticletitle{The Synthetic Data Challenge}.
\newblock \bibinfo{journal}{\emph{Conference of European Statisticians}}
  (\bibinfo{year}{2019}).
\newblock
\urldef\tempurl%
\url{https://unece.org/fileadmin/DAM/stats/documents/ece/ces/ge.46/2019/mtg1/SDC2019_S3_UK_Synthethic_Data_Challenge_Elliot_AD.pdf}
\showURL{%
\tempurl}


\bibitem[\protect\citeauthoryear{Karr, Kohnen, Oganian, Reiter, and Sanil}{Karr
  et~al\mbox{.}}{2006}]%
        {cio}
\bibfield{author}{\bibinfo{person}{A.~F Karr}, \bibinfo{person}{C.~N Kohnen},
  \bibinfo{person}{A Oganian}, \bibinfo{person}{J.~P Reiter}, {and}
  \bibinfo{person}{A.~P Sanil}.} \bibinfo{year}{2006}\natexlab{}.
\newblock \showarticletitle{A Framework for Evaluating the Utility of Data
  Altered to Protect Confidentiality}.
\newblock \bibinfo{journal}{\emph{The American Statistician}}
  \bibinfo{volume}{60}, \bibinfo{number}{3} (\bibinfo{year}{2006}),
  \bibinfo{pages}{224--232}.
\newblock
\urldef\tempurl%
\url{https://doi.org/10.1198/000313006X124640}
\showDOI{\tempurl}
\showeprint{https://doi.org/10.1198/000313006X124640}


\bibitem[\protect\citeauthoryear{Lin, Khetan, Fanti, and Oh}{Lin
  et~al\mbox{.}}{2020}]%
        {pacgan}
\bibfield{author}{\bibinfo{person}{Zinan Lin}, \bibinfo{person}{Ashish Khetan},
  \bibinfo{person}{Giulia Fanti}, {and} \bibinfo{person}{Sewoong Oh}.}
  \bibinfo{year}{2020}\natexlab{}.
\newblock \showarticletitle{PacGAN: The Power of Two Samples in Generative
  Adversarial Networks}.
\newblock \bibinfo{journal}{\emph{IEEE Journal on Selected Areas in Information
  Theory}} \bibinfo{volume}{1}, \bibinfo{number}{1} (\bibinfo{year}{2020}),
  \bibinfo{pages}{324--335}.
\newblock
\urldef\tempurl%
\url{https://doi.org/10.1109/JSAIT.2020.2983071}
\showDOI{\tempurl}


\bibitem[\protect\citeauthoryear{Little, Elliot, and Allmendinger}{Little
  et~al\mbox{.}}{2022}]%
        {framework}
\bibfield{author}{\bibinfo{person}{Claire Little}, \bibinfo{person}{Mark
  Elliot}, {and} \bibinfo{person}{Richard Allmendinger}.}
  \bibinfo{year}{2022}\natexlab{}.
\newblock \showarticletitle{Comparing the Utility and Disclosure Risk of
  Synthetic Data with Samples of Microdata}. In
  \bibinfo{booktitle}{\emph{Privacy in Statistical Databases}},
  \bibfield{editor}{\bibinfo{person}{Josep Domingo-Ferrer} {and}
  \bibinfo{person}{Maryline Laurent}} (Eds.). \bibinfo{publisher}{Springer
  International Publishing}, \bibinfo{address}{Cham},
  \bibinfo{pages}{234--249}.
\newblock
\showISBNx{978-3-031-13945-1}


\bibitem[\protect\citeauthoryear{Little, Elliot, and Allmendinger}{Little
  et~al\mbox{.}}{2023}]%
        {little2023federated}
\bibfield{author}{\bibinfo{person}{Claire Little}, \bibinfo{person}{Mark
  Elliot}, {and} \bibinfo{person}{Richard Allmendinger}.}
  \bibinfo{year}{2023}\natexlab{}.
\newblock \showarticletitle{Federated learning for generating synthetic data: a
  scoping review}.
\newblock \bibinfo{journal}{\emph{International Journal of Population Data
  Science}} \bibinfo{volume}{8}, \bibinfo{number}{1} (\bibinfo{year}{2023}),
  \bibinfo{pages}{2158}.
\newblock
\urldef\tempurl%
\url{https://doi.org/10.23889/ijpds.v8i1.2158}
\showDOI{\tempurl}


\bibitem[\protect\citeauthoryear{Little, Elliot, and Allmendinger}{Little
  et~al\mbox{.}}{2024}]%
        {little2024benchmarking}
\bibfield{author}{\bibinfo{person}{Claire Little}, \bibinfo{person}{Mark
  Elliot}, {and} \bibinfo{person}{Richard Allmendinger}.}
  \bibinfo{year}{2024}\natexlab{}.
\newblock \showarticletitle{Synthetic census microdata generation: A
  comparative study of synthesis methods examining the trade-off between
  disclosure risk and utility}.
\newblock \bibinfo{journal}{\emph{Journal of Official Statistics}}
  (\bibinfo{year}{2024}).
\newblock


\bibitem[\protect\citeauthoryear{Ma, Tschiatschek, Turner,
  Hern\'{a}ndez-Lobato, and Zhang}{Ma et~al\mbox{.}}{2020}]%
        {ma2020vaem}
\bibfield{author}{\bibinfo{person}{Chao Ma}, \bibinfo{person}{Sebastian
  Tschiatschek}, \bibinfo{person}{Richard Turner},
  \bibinfo{person}{Jos\'{e}~Miguel Hern\'{a}ndez-Lobato}, {and}
  \bibinfo{person}{Cheng Zhang}.} \bibinfo{year}{2020}\natexlab{}.
\newblock \showarticletitle{VAEM: A Deep Generative Model for Heterogeneous
  Mixed Type Data}. In \bibinfo{booktitle}{\emph{Proceedings of the 34th
  International Conference on Neural Information Processing Systems}}
  (Vancouver, BC, Canada) \emph{(\bibinfo{series}{NIPS'20})}.
  \bibinfo{publisher}{Curran Associates Inc.}, \bibinfo{address}{Red Hook, NY,
  USA}, Article \bibinfo{articleno}{943}, \bibinfo{numpages}{11}~pages.
\newblock
\showISBNx{9781713829546}


\bibitem[\protect\citeauthoryear{Makhzani, Shlens, Jaitly, and
  Goodfellow}{Makhzani et~al\mbox{.}}{2016}]%
        {makhzani2016adversarial}
\bibfield{author}{\bibinfo{person}{Alireza Makhzani}, \bibinfo{person}{Jonathon
  Shlens}, \bibinfo{person}{Navdeep Jaitly}, {and} \bibinfo{person}{Ian
  Goodfellow}.} \bibinfo{year}{2016}\natexlab{}.
\newblock \showarticletitle{Adversarial Autoencoders}. In
  \bibinfo{booktitle}{\emph{International Conference on Learning
  Representations}}.
\newblock
\urldef\tempurl%
\url{http://arxiv.org/abs/1511.05644}
\showURL{%
\tempurl}


\bibitem[\protect\citeauthoryear{Mao, Li, Xie, Lau, Wang, and Paul~Smolley}{Mao
  et~al\mbox{.}}{2017}]%
        {lsgan}
\bibfield{author}{\bibinfo{person}{Xudong Mao}, \bibinfo{person}{Qing Li},
  \bibinfo{person}{Haoran Xie}, \bibinfo{person}{Raymond~Y.K. Lau},
  \bibinfo{person}{Zhen Wang}, {and} \bibinfo{person}{Stephen Paul~Smolley}.}
  \bibinfo{year}{2017}\natexlab{}.
\newblock \showarticletitle{Least Squares Generative Adversarial Networks}. In
  \bibinfo{booktitle}{\emph{Proceedings of the IEEE International Conference on
  Computer Vision (ICCV)}}.
\newblock


\bibitem[\protect\citeauthoryear{Nasution, Bhaswara, Nugraha, and
  Kanggrawan}{Nasution et~al\mbox{.}}{2022}]%
        {nasution2022data}
\bibfield{author}{\bibinfo{person}{Bahrul~Ilmi Nasution},
  \bibinfo{person}{Irfan~Dwiki Bhaswara}, \bibinfo{person}{Yudhistira Nugraha},
  {and} \bibinfo{person}{Juan~Intan Kanggrawan}.}
  \bibinfo{year}{2022}\natexlab{}.
\newblock \showarticletitle{Data Analysis and Synthesis of COVID-19 Patients
  using Deep Generative Models: A Case Study of Jakarta, Indonesia}.
\newblock \bibinfo{journal}{\emph{2022 IEEE International Smart Cities
  Conference (ISC2)}}, \bibinfo{pages}{1--7}.
\newblock
\urldef\tempurl%
\url{https://doi.org/10.1109/ISC255366.2022.9921948}
\showDOI{\tempurl}


\bibitem[\protect\citeauthoryear{Nouri, Ghandri, Driss, and Ghedira}{Nouri
  et~al\mbox{.}}{2023}]%
        {Nouri2023bievogan}
\bibfield{author}{\bibinfo{person}{Houssem~Eddine Nouri},
  \bibinfo{person}{Abdennaceur Ghandri}, \bibinfo{person}{Olfa~Belkahla Driss},
  {and} \bibinfo{person}{Khaled Ghedira}.} \bibinfo{year}{2023}\natexlab{}.
\newblock \showarticletitle{Bi-{EvoGAN}: Bi-level Evolutionary Approach for
  Generative Adversarial Networks}.
\newblock \bibinfo{journal}{\emph{Applied Soft Computing}}
  \bibinfo{volume}{147} (\bibinfo{date}{Nov.} \bibinfo{year}{2023}),
  \bibinfo{pages}{110738}.
\newblock
\urldef\tempurl%
\url{https://doi.org/10.1016/j.asoc.2023.110738}
\showDOI{\tempurl}


\bibitem[\protect\citeauthoryear{Nowok, Raab, and Dibben}{Nowok
  et~al\mbox{.}}{2016}]%
        {synthpop}
\bibfield{author}{\bibinfo{person}{Beata Nowok}, \bibinfo{person}{Gillian~M.
  Raab}, {and} \bibinfo{person}{Chris Dibben}.}
  \bibinfo{year}{2016}\natexlab{}.
\newblock \showarticletitle{synthpop: Bespoke Creation of Synthetic Data in R}.
\newblock \bibinfo{journal}{\emph{Journal of Statistical Software}}
  \bibinfo{volume}{74} (\bibinfo{year}{2016}), \bibinfo{pages}{1–26}.
\newblock


\bibitem[\protect\citeauthoryear{Nowozin, Cseke, and Tomioka}{Nowozin
  et~al\mbox{.}}{2016}]%
        {nowozin2016fgan}
\bibfield{author}{\bibinfo{person}{Sebastian Nowozin}, \bibinfo{person}{Botond
  Cseke}, {and} \bibinfo{person}{Ryota Tomioka}.}
  \bibinfo{year}{2016}\natexlab{}.
\newblock \showarticletitle{F-GAN: Training Generative Neural Samplers Using
  Variational Divergence Minimization}. In
  \bibinfo{booktitle}{\emph{Proceedings of the 30th International Conference on
  Neural Information Processing Systems}} (Barcelona, Spain)
  \emph{(\bibinfo{series}{NIPS'16})}. \bibinfo{publisher}{Curran Associates
  Inc.}, \bibinfo{address}{Red Hook, NY, USA}, \bibinfo{pages}{271–279}.
\newblock
\showISBNx{9781510838819}


\bibitem[\protect\citeauthoryear{{Office for National Statistics, Census
  Division} and {University of Manchester, Cathie Marsh Centre for Census and
  Survey Research}}{{Office for National Statistics, Census Division} and
  {University of Manchester, Cathie Marsh Centre for Census and Survey
  Research}}{2013}]%
        {uk_data}
\bibfield{author}{\bibinfo{person}{{Office for National Statistics, Census
  Division}} {and} \bibinfo{person}{{University of Manchester, Cathie Marsh
  Centre for Census and Survey Research}}.} \bibinfo{year}{2013}\natexlab{}.
\newblock \showarticletitle{Census 1991: Individual Sample of Anonymised
  Records for Great Britain (SARs)}.
\newblock  (\bibinfo{year}{2013}).
\newblock


\bibitem[\protect\citeauthoryear{Ping, Stoyanovich, and Howe}{Ping
  et~al\mbox{.}}{2017}]%
        {data_synthesizer}
\bibfield{author}{\bibinfo{person}{Haoyue Ping}, \bibinfo{person}{Julia
  Stoyanovich}, {and} \bibinfo{person}{Bill Howe}.}
  \bibinfo{year}{2017}\natexlab{}.
\newblock \showarticletitle{DataSynthesizer: Privacy-Preserving Synthetic
  Datasets}. In \bibinfo{booktitle}{\emph{Proceedings of the 29th International
  Conference on Scientific and Statistical Database Management}} (Chicago, IL,
  USA) \emph{(\bibinfo{series}{SSDBM '17})}. \bibinfo{publisher}{Association
  for Computing Machinery}, \bibinfo{address}{New York, NY, USA}, Article
  \bibinfo{articleno}{42}, \bibinfo{numpages}{5}~pages.
\newblock
\showISBNx{9781450352826}
\urldef\tempurl%
\url{https://doi.org/10.1145/3085504.3091117}
\showDOI{\tempurl}


\bibitem[\protect\citeauthoryear{Purdam and Elliot}{Purdam and Elliot}{2007}]%
        {Purdam2007}
\bibfield{author}{\bibinfo{person}{Kingsley Purdam} {and} \bibinfo{person}{Mark
  Elliot}.} \bibinfo{year}{2007}\natexlab{}.
\newblock \showarticletitle{A case study of the impact of statistical
  disclosure control on data quality in the individual UK Samples of Anonymised
  Records.}
\newblock \bibinfo{journal}{\emph{Environment and Planning A}}
  \bibinfo{volume}{39}, \bibinfo{number}{5} (\bibinfo{year}{2007}),
  \bibinfo{pages}{1101--1118}.
\newblock


\bibitem[\protect\citeauthoryear{Radford, Metz, and Chintala}{Radford
  et~al\mbox{.}}{2015}]%
        {dcgan}
\bibfield{author}{\bibinfo{person}{Alec Radford}, \bibinfo{person}{Luke Metz},
  {and} \bibinfo{person}{Soumith Chintala}.} \bibinfo{year}{2015}\natexlab{}.
\newblock \bibinfo{title}{Unsupervised Representation Learning with Deep
  Convolutional Generative Adversarial Networks}.
\newblock
\newblock
\urldef\tempurl%
\url{https://doi.org/10.48550/ARXIV.1511.06434}
\showDOI{\tempurl}


\bibitem[\protect\citeauthoryear{Saatci and Wilson}{Saatci and Wilson}{2017}]%
        {saatci2017bayesian}
\bibfield{author}{\bibinfo{person}{Yunus Saatci} {and}
  \bibinfo{person}{Andrew~G Wilson}.} \bibinfo{year}{2017}\natexlab{}.
\newblock \showarticletitle{Bayesian GAN},
  \bibfield{editor}{\bibinfo{person}{I~Guyon}, \bibinfo{person}{U~Von Luxburg},
  \bibinfo{person}{S~Bengio}, \bibinfo{person}{H~Wallach},
  \bibinfo{person}{R~Fergus}, \bibinfo{person}{S~Vishwanathan}, {and}
  \bibinfo{person}{R~Garnett}} (Eds.).
\newblock \bibinfo{journal}{\emph{Advances in Neural Information Processing
  Systems}}  \bibinfo{volume}{30}.
\newblock
\urldef\tempurl%
\url{https://proceedings.neurips.cc/paper/2017/file/312351bff07989769097660a56395065-Paper.pdf}
\showURL{%
\tempurl}


\bibitem[\protect\citeauthoryear{Sutton and Barto}{Sutton and Barto}{2018}]%
        {sutton2018reinforcement}
\bibfield{author}{\bibinfo{person}{Richard~S Sutton} {and}
  \bibinfo{person}{Andrew~G Barto}.} \bibinfo{year}{2018}\natexlab{}.
\newblock \bibinfo{booktitle}{\emph{Reinforcement learning: An introduction}}.
\newblock \bibinfo{publisher}{MIT press}.
\newblock


\bibitem[\protect\citeauthoryear{Tomczak}{Tomczak}{2022}]%
        {Tomczak2022deep}
\bibfield{author}{\bibinfo{person}{Jakub~M. Tomczak}.}
  \bibinfo{year}{2022}\natexlab{}.
\newblock \bibinfo{booktitle}{\emph{Deep Generative Modeling}}.
\newblock \bibinfo{publisher}{Springer International Publishing}.
\newblock
\showISBNx{978-3-030-93157-5}
\urldef\tempurl%
\url{https://doi.org/10.1007/978-3-030-93158-2}
\showDOI{\tempurl}


\bibitem[\protect\citeauthoryear{Turner, Hung, Frank, Saatci, and
  Yosinski}{Turner et~al\mbox{.}}{2018}]%
        {turner2018metropolis}
\bibfield{author}{\bibinfo{person}{Ryan Turner}, \bibinfo{person}{Jane Hung},
  \bibinfo{person}{Eric Frank}, \bibinfo{person}{Yunus Saatci}, {and}
  \bibinfo{person}{Jason Yosinski}.} \bibinfo{year}{2018}\natexlab{}.
\newblock \showarticletitle{Metropolis-Hastings Generative Adversarial
  Networks}.
\newblock  (\bibinfo{date}{11} \bibinfo{year}{2018}).
\newblock
\urldef\tempurl%
\url{https://doi.org/10.48550/arxiv.1811.11357}
\showDOI{\tempurl}


\bibitem[\protect\citeauthoryear{Wang, Xu, Yao, and Tao}{Wang
  et~al\mbox{.}}{2019}]%
        {egan}
\bibfield{author}{\bibinfo{person}{Chaoyue Wang}, \bibinfo{person}{Chang Xu},
  \bibinfo{person}{Xin Yao}, {and} \bibinfo{person}{Dacheng Tao}.}
  \bibinfo{year}{2019}\natexlab{}.
\newblock \showarticletitle{Evolutionary Generative Adversarial Networks}.
\newblock \bibinfo{journal}{\emph{IEEE Transactions on Evolutionary
  Computation}} \bibinfo{volume}{23}, \bibinfo{number}{6}
  (\bibinfo{year}{2019}), \bibinfo{pages}{921--934}.
\newblock
\urldef\tempurl%
\url{https://doi.org/10.1109/TEVC.2019.2895748}
\showDOI{\tempurl}


\bibitem[\protect\citeauthoryear{Wang, Zheng, He, Chen, and Zhou}{Wang
  et~al\mbox{.}}{2023}]%
        {wang2023diffusiongan}
\bibfield{author}{\bibinfo{person}{Zhendong Wang}, \bibinfo{person}{Huangjie
  Zheng}, \bibinfo{person}{Pengcheng He}, \bibinfo{person}{Weizhu Chen}, {and}
  \bibinfo{person}{Mingyuan Zhou}.} \bibinfo{year}{2023}\natexlab{}.
\newblock \bibinfo{title}{Diffusion-GAN: Training GANs with Diffusion}.
\newblock
\newblock
\showeprint[arxiv]{2206.02262}~[cs.LG]


\bibitem[\protect\citeauthoryear{Xu, Skoularidou, Cuesta-Infante, and
  Veeramachaneni}{Xu et~al\mbox{.}}{2019}]%
        {ctgan}
\bibfield{author}{\bibinfo{person}{Lei Xu}, \bibinfo{person}{Maria
  Skoularidou}, \bibinfo{person}{Alfredo Cuesta-Infante}, {and}
  \bibinfo{person}{Kalyan Veeramachaneni}.} \bibinfo{year}{2019}\natexlab{}.
\newblock \showarticletitle{Modeling Tabular data using Conditional GAN}. In
  \bibinfo{booktitle}{\emph{Advances in Neural Information Processing
  Systems}}, \bibfield{editor}{\bibinfo{person}{H.~Wallach},
  \bibinfo{person}{H.~Larochelle}, \bibinfo{person}{A.~Beygelzimer},
  \bibinfo{person}{F.~d\textquotesingle Alch\'{e}-Buc},
  \bibinfo{person}{E.~Fox}, {and} \bibinfo{person}{R.~Garnett}} (Eds.),
  Vol.~\bibinfo{volume}{32}. \bibinfo{publisher}{Curran Associates, Inc.}
\newblock
\urldef\tempurl%
\url{https://proceedings.neurips.cc/paper/2019/file/254ed7d2de3b23ab10936522dd547b78-Paper.pdf}
\showURL{%
\tempurl}


\bibitem[\protect\citeauthoryear{Zhou, Yao, Yue, and Niu}{Zhou
  et~al\mbox{.}}{2023}]%
        {Zhou2023wgan}
\bibfield{author}{\bibinfo{person}{Tianwei Zhou}, \bibinfo{person}{Xizhang
  Yao}, \bibinfo{person}{Guanghui Yue}, {and} \bibinfo{person}{Ben Niu}.}
  \bibinfo{year}{2023}\natexlab{}.
\newblock \showarticletitle{A {WGAN}-Based Generative Strategy in Evolutionary
  Multitasking for Multi-objective Optimization}.
\newblock In \bibinfo{booktitle}{\emph{Lecture Notes in Computer Science}}.
  \bibinfo{publisher}{Springer Nature Switzerland}, \bibinfo{pages}{390--400}.
\newblock
\urldef\tempurl%
\url{https://doi.org/10.1007/978-3-031-36622-2_32}
\showDOI{\tempurl}


\end{thebibliography}

\appendix

\end{document}